# Problem Solving and Computers in a Learning Environment

**Michael Gr. Voskoglou [1]**, **Sheryl Buckley [2]**

[1] School of Technological Applications
Graduate Technological Educational Institute of Patras, GREECE
mvosk@hol.gr   , voskoglou@teipat.gr
[2] School of Computing (Acting)
University of South Africa, Florida Campus, SOUTH AFRICA
bucklsb@unisa.ac.za

## Abstract

Computational thinking is a new problem solving method named for its extensive use of computer science techniques. It synthesizes critical thinking and existing knowledge and applies them to solve complex technological problems. The term was coined by J. Wing [1], but the relationship between computational and critical thinking, the two modes of thinking in solving problems, has not been yet clearly established. This paper aims in shedding some light into this relationship. We also present two classroom experiments performed recently at the Graduate Technological Educational Institute (TEI) of Patras, Greece. The result of these experiment give a strong indication that the use of computers as a tool for problem solving enhances the students' abilities in solving real world problems involving mathematical modelling. This is crossed by earlier findings of other researchers for the problem solving process in general (not only for mathematical problems).

## 1. Introduction

The importance of Problem Solving (PS) has been realised for such a long time that in a direct or indirect way affects our daily lives for decades. Volumes of research have been written about PS and attempts have been made by many educationists and psychologists to make it accessible to all in various degrees. No wonder it is entrenched in educational policies of most countries, e.g. American college personnel association,1994, cited by King & Baxter-Magolda [2]; South African Qualifications Authority [3], etc.  And wherever numerical problems are involved from the simple 2 + 3 to complex numerical analysis, technological tools (from a simple calculator to sophisticated computers) have been developed to assist the problem solver to solve the problem effectively and efficiently.  Gone are the logarithmic tables and slide rules.

However, even graduates have nowadays difficulty in solving real life problems. Somehow, they can not apply theory into practice, or theorise/reflect on practice ([4], [5], [6]). In fact, it is the human mind in the end that has to be applied in a problematic situation and solve the problem.  Its capacity to solve the problem is directly related to the knowledge stored in the mind.  And knowledge is the product of thinking.  But thinking can vary from a very simple and mundane thought to a very sophisticated and complex one [7].  The nature of the problem dictates the level of thinking.

Higher-order thinking can be conceptualised as a complex mode of thinking that often generates multiple solutions [8].  Such thinking involves abstraction, uncertainty, application of multiple criteria, reflection, and self-regulation. It also facilitates the transfer of knowledge, i.e. the use and transformation of already existing knowledge in creating new knowledge (e.g. see [9], section 3). On the other hand, lower-order thinking could be considered to be one that requires minimum cognitive effort and it is algorithmic.  In an attempt by humans to increase the power of the mind, Herbet Simon saw thinking as information-processing [10] and computers started taking over as a kind of 'thinking machines'.  McGuinness [10] also saw thinking as making judgements and sense-making.  And making judgements is directly related to *critical thinking*.

The complexity of critical thinking is evident from the fact that there is no definition that is universally accepted.  However, a great number of critical thinking skills as identified by are agreed upon by many authors.  Some of these skills are: analysis and synthesis, making judgements, decision making, drawing warranted conclusions and generalisations.  Critical thinking is thus a prerequisite to PS.

However, when computers are used in the PS situation, the need for *computational thinking (CT)* is another prerequisite.  CT has been coined by J. Wing [1] and broadly speaking it describes a set of thinking skills that are integral to solving complex problems using a computer.

Living in a knowledge era and an ever increasing progress in technology, combining knowledge and technology to solve problems is becoming the mode rather than the exception. Creativity and innovation driven by tacit knowledge, and critical thinking driven by logic, making judgements, analysis and synthesis and so on become the tools for problem thinking and PS. If technology is added as another tool then CT (in its broader meaning other than performing computations) is a prerequisite.  For Halpern [11] and Matlin 12] it frees memory so that the problem solver can concentrate on the essence of the problem.

The relationship between CT and critical thinking, the two modes of thinking in solving problems, has not yet been clearly established.  This paper aims at shedding some light into this relationship.

## 2. The PS process: A review

It is agreed by many authors ([11], [4], [5]) that PS is a complex phenomenon and no wonder there is no unique definition. However, the following definitions encompass most of the existing definitions:

For Polya [13], the pioneer in PS, "solving a problem means finding a way out of a difficulty, a way around an obstacle, attaining an aim that was not immediately understandable." According to Schoenfeld [14] "a problem is only a problem, if you don't know how to go about solving it. A problem that has no 'surprises' in store, and can be solved comfortably by routine or familiar procedures (no matter how difficult!) it is an exercise." Green and Gilhooly [5] state that "PS in all its manifestations is an activity that structures everyday life in a meaningful way." The authors add further that this activity draws together the different components of cognition. Therefore, the kind of problem will dictate the type of cognitive skill necessary to solve the problem: linguistic skills are used to read about a certain problem and debate about it, memory skills to recall prior knowledge and so on. Depending on the knowledge and thinking skills possessed by a problem solver, what could be a problem for one might not be a problem for some body else. Perhaps Martinez's [4] definition carries the modern message about PS: "PS can be defined simply as the pursuit of a goal when the path to that goal is uncertain. In other words, it's what you do when you don't know what you're doing."

In concluding, we adopt the following working definition for this paper: *PS is an activity that makes use of cognitive or cognitive and physical means to overcome an obstacle (problem) and develop a better idea of the world that surrounds us.*

Most authors agree that PS is a cognitive process, though according to the latest findings [15] it could be considered rather as the product of a number of processes (cognitive actions), since it is not actually the process/processes that are of value, but the successful solution of the problem. Engaging in PS implies conscious and subconscious thinking. And the type of problem will dictate the type of thinking: The more complex the problem, the higher the level of thinking required.

Mathematics by its nature is a subject whereby PS forms its essence. In an earlier paper [9] we have examined the role of the problem in learning mathematics and we have attempted a review of the evolution of research on PS in mathematics education from its emergency as a self sufficient science at the 1960's until today. Here is a rough chronology of that progress:
1950's – 1960's: Polya's ideas on the use of heuristic strategies in PS ([13], [16], etc).

1970's: Emergency of mathematics education as a self − sufficient science (research methods were almost exclusively statistical). Research on PS was still based on Polya's ideas.

1980's: A framework describing the PS process, and reasons for success or failure in PS: 'Expert performance model' ([17], [18], etc).

1990's: Models of teaching using PS, e.g. constructivist view of learning [19], mathematical modelling and applications (e.g. see [20] and its references), etc.

2000's: While early work on PS focused mainly on analyzing the PS process and on describing the proper heuristic strategies to be used in each of its stages, more recent investigations have focused mainly on solvers' behaviour and required attributes during the PS process; e.g. 'Multidimensional PS Framework' (MPSF) of Carlson & Bloom [21]. More comprehensive models for the PS process in general (not only for mathematics) were developed by Sternberg & Ben-Zeev [22],  by Schoenfeld  [23] ('PS as a goal-oriented behaviour'), etc.

What has been agreed by many authors ([5], [11], [12], etc) is that a problem basically consists of three states: the *starting state,* the *goal state* and the *obstacles* or a set of available actions or strategies to move from the starting state to the goal state. Matlin [12] states that the initial state describes the situation at the beginning of the problem; the goal state is reached when we solved the problem; and the obstacles describe the restrictions that make it difficult to proceed from the initial state to the goal state.  *The greatest difference among problem solvers of the same problem tends to lie in the third state, which might have 'infinite' possibilities, if the goal and the starting points are the same for every one.*

## 3. Critical thinking and CT in PS

Critical thinking has its roots in the ancient Greek philosopher, Socrates, who perfected the art of questioning whereby by asking pertinent questions he would show that "people could not rationally justify their confident claims of knowledge".

The importance of critical thinking is one of the seven educational critical outcomes together with PS and many authors support this ([11], [24], [25], [4], [7], [26], etc). With the explosion of information technology and moving away from an industrial society to a knowledge society, the attitude or disposition to think critically is as important as other skills such as professional acquisition of knowledge and lifelong learning ([27], [11], [24]). Understanding what critical thinking is, how it can be acquired might help institutions of learning to instil such skills in the learner and become more effective and efficient.

Halpern's [11] idea of thinking as *information processing* is complemented by McGuinness's [10] idea of thinking as *making judgements and sense-making*.  The former, deals with critical thinking, while the latter is embedded in constructivism.  Critical thinking is a particular human activity and a complex way of thinking.  For this study a combination of the definitions by Halpern [11] and Williams [26]  gave rise into the following working definition:

*Critical thinking is an ability or skill by which the individual transcends his/her subjective self in a wilful manner in order to arrive rationally at conclusions (not necessarily favourable to him/ her) that can be substantiated using valid information.*

Critical thinking is considered to be a higher, non algorithmic, complex mode of thinking that often generates multiple solutions. Through it thinking skills of higher level such as analysis, synthesis and evaluation are combined giving rise to other skills like inferring, estimating, predicting, generalising, and creative thinking and PS. Therefore, PS is preceded by critical thinking.

But critical thinking also affects acquisition of knowledge as knowledge is the product of thinking about concepts and combining them with principles. Concepts are acquired through abstractions and principles connect the concepts thus forming a network.  Any new concept encountered has to fit in the existing cognitive structure.  Such accommodation will not be possible without critical thinking.  When a problem is encountered, before being solved it has to be analysed in a critical way: What is the problem, what is the given information and so on. Therefore, critical thinking is also involved in application of knowledge to solve the problem.

It can be concluded that critical thinking is a prerequisite to knowledge acquisition and application to solve problems, but not a sufficient condition when we are faced with complex real technological problems. Technological problems require also a pragmatic way of thinking such as CT.

Computation is an increasingly essential tool for doing scientific research.  It is expected that future generations of engineers will need to engage and understand computing in order to work effectively with computational systems, technologies and methodologies.  CT is a type of analytical thinking that employs mathematical and engineering thinking to understand and solve complex problems within the constraints of the real world. The term was first used by S. Papert [28], who is widely known for the development of the Logo software. However, it was brought to the forefront of the computer society by Wing [1] to describe how to think like a computer scientist.  She described CT as *"solving problems, designing systems and understanding human behaviour by drawing on the concepts fundamental to computer science."*

The main characteristics of CT include:
- Analyzing and logically organizing data
- Data modelling, data abstractions, and simulations
- Formulating problems such that computers may assist
- Identifying, testing, and implementing possible solutions
- Automating solutions via algorithmic thinking
- Generalizing and applying this process to other problems
  ( http://en.wikipedia.org/wiki/Computational_thinking )

Thus, according to Liu and Wang [29] *computational thinking is a hybrid of other modes of thinking*, like abstract thinking, logical thinking, modelling thinking, and constructive thinking:

In order to understand the main body of computer problem, *abstract thinking* is essential in computer science and technology.  In solving an interesting problem, abstraction of thinking is one very general purpose heuristic that can help to attack this problem.

Informally, abstraction thinking can be thought of the mapping from a ground representation to a new but simpler representation.

*Logical thinking* is the process in which one uses reasoning consistency to come to a conclusion. Some computer problems or computer states (situations) involving logical thinking always call for mathematics structure, for relationships between some hypotheses and given statements, and for a sequence of reasoning that makes the conclusion more reasonable.

*Modelling thinking*, in the technical use of the term, refers to the translation of objects or phenomena from the real world into mathematical equations (mathematical models) or computer relations (simulation models). It is choosing an appropriate representation or modelling the relevant aspects of a problem to make it tractable. Computer modelling is the representation of reality objects on a computer. A problem which will be solved by computer must be modelled by a corresponding software model.

*Constructive thinking* is any well-defined computational procedure that takes some value, or set of values as input and produces some value, or set of values as output.

What is evident from the above discussion is that when we have to solve real complex technological problems computational thinking with its components as described by Liu and Wang [29] are necessary thinking modes. *They synthesise critical thinking and existing knowledge and apply them to solve the problem.* CT does not propose that problems need to be solved in the same way a computer tackles them, *but rather it encourages the critical thinking using computer science concepts and techniques*. Thus CT is a prerequisite to PS when we face real complex technological problems.

Among computer science articles ([30], [31]), the characteristics most commonly referred to about CT are abstraction and PS. The ability to think of problems in a more abstract manner is essential to students' ability to see more solution opportunities. Thinking in this abstract manner, or computationally, is a way of accessing solutions that are usually outside a student's normal area of expertise or schemata. According to Anderson et al. [32], "to interpret a particular situation in terms of a schema is to match the elements in the situation with the generic characterisations in the schematic knowledge structure." Learning to think computationally or to problem-solve through abstraction is the ability to eliminate details from a given situation in order to find a solution that might not be forthcoming under other circumstances [33].

CT can involve solving mathematical problems, building engineering systems, interpreting data, etc. It is becoming recognised as an important way to educate new generations of students who will become skilled not only at using tools, but also at creating them. All of today's students will go on to live a life heavily influenced by computing, and many will work in fields that involve or are influenced by computing [34]. There is therefore a need to start teaching CT early and often [35].

CT today is spearheaded by the Center of Computational Thinking at Carnegie Mellon University in Pittsburg, Pensylvania, USA. The Center's major activity is conducting PROBEs (**PROB**lem-oriented **E**xplorations). These PROBEs are experiments that apply novel computing concepts to problems to show the value of CT. A PROBE experiment is generally a collaboration between a computer scientist and an expert in the field to be studied. The experiment typically runs for a year. In general, a PROBE will seek to find a solution for a broadly applicable problem and avoid narrowly focused issues. Some examples of PROBE experiments are optimal kidney transplant logistics and how to create drugs that do not breed drug resistant viruses (for more details look at www.cs.cmu.edu/~CompThink/probes.html).

According to Denning [36] CT is just a new term for a core concept, algorithmic thinking.  It is the new term for the more user-friendly Computer Science discipline. Computer Science is not just about programming, it's about an entire way of thinking, which is now an intrinsic part of our lives. One could argue that a world without computers would be unthinkable.

Yet there is a consensus that computer science has serious conundrums, such as attracting students, low retention rates and low motivation for learning programming despite the continuing growth of the information technology industry [37].  It is widely accepted that motivation and involvement are imperative in retaining students in computer science and in order to do this we need to engage students more in the process of learning programming by building more effective mechanisms and tools for the development of programming skills. However, this is not an easy task and one of the core aims of learning programming should be to constantly highlight that programming is not only coding, but also thinking computationally and acquiring skills to develop solid solutions through understanding of concrete problems. Recent studies in this field address the necessity *to become trained in thinking computationally before learning programming*, and conclude that the education of programming along with the theory of computing needs to be represented in a way that would make sense to students within the computer science discipline [37].

In thinking as a computer scientist, researchers become aware of behaviours and reactions that can be captured in algorithms or can be analysed within an algorithmic framework.  CT now gives them a different framework for visualizing and analysing, a whole new perspective.  To rephrase a common idiom, "until you have a screwdriver, everything looks like a nail."  CT develops a variety of skills (logic, creativity, algorithmic thinking, modelling/simulations), involves the use of scientific methodologies and helps developing both inventiveness and innovative thinking.  It has roots in mathematics, engineering, technology and science and in the synthesis of ideas from all these fields, has created a way of thinking that is only just beginning to generate enormous changes and benefits [38].

CT is a learned approach and there's no better way to learn it explicitly than through programming.  Programming employs all the components of CT and the knowledge gained through the experience of tackling programming challenges – both explicit and tacit – can provide a framework not only for computer science, but for any field from natural and health sciences to the social sciences and humanities.

So, here we have an important, essential and very truly 21$^{st}$ century skill, CT that is best learned through experience, interactions, and actively doing.  It allows students who learn to express themselves through programming (and who have the time to gain this knowledge) to not only answer questions, but also generate new ones as they begin to view these challenges through the lens of the tacit knowledge intrinsic to computational thinking.  The art of programming requires creativity and inventiveness, logic, algorithmic thinking and an appreciation of the recursive nature of this process, as the student learns from her failures, refines his/her work and gets a deeper understanding of the problem.  As with any creation, even once a solution is found – a pattern, an algorithm – the solution can be refined, simplified and beautified, made more elegant.  In a way, programming provides the same satisfaction as a video game – the opportunity to find a path – one of many – through a problem.  The difference here is that students can answer their own questions and create their own challenges [38].

CT is the computer science approach to PS through abstraction and innovation, through abstractly conceptualising the question or challenge [39].  And the more frequently student's use CT, the better they will become at finding alternative, unique and inventive solutions to complex problems.

Our exploration into the dependence of PS of real engineering and scientific problems on CT and critical thinking revealed that there is a strong link between the constructs.  As a result teaching practices should take cognisance of this finding.

## 4. The Computers as a Tool for PS

The role that the rational use of the new technologies could play for the development of students' PS abilities is very important indeed. In fact, the animation of figures and mathematical representations, provided by suitable computer software packages, videos, etc, increases the students' imagination and helps them in finding solutions easier of the corresponding problems. The role of mathematical theory after this is not to convince, but to explain [40].

The type of the problem, as well as the level of human development (cognitive, physical) will dictate the type of skills necessary to solve that problem. Certain problems, and especially the complex technological problems of real life, require more than knowledge and logical analysis (Brookfield 1987).  They require complex cognitive skills and possibly sophisticated available tools such as computers.  And computers are viewed by some as 'tools that can perform miracles', solve any problem, while by others as auxiliary tools which perform cumbersome operations and use the saved time for functions that a computer cannot perform, i.e. creative thinking, developing new ideas and so on [38].

While the former group might have an exaggerated view as to what computers do, the latter, depending on the degree of computer science knowledge, might underestimate the power of the computer.  The fact is that they both know that a computer (the hardware) was created by a human being, and comes into 'life' through programming, which was also done by a human being.  And through programming it is possible to input information and get an output almost at the speed of light. The old credo though "garbage in, garbage out" is still valid.

The power of the computer as an information processing device was used to develop information processing technologies. Indeed Halpern [11] sees thinking as human information processing (a cognitive perspective), and the brain like a computer with a processor of "infinite memory."  This idea was used by instructional designers to supplement learning and teach students to practice better ways of thinking.  Technologies for face to face as well as for distant education developed in leaps and bounds.  Reality is now experienced also from a distance as a virtual reality.

According to information-processing psychology, it seeks to study the mind in general, and intelligence, in particular, in terms of mental representations and processes that underlie observable behaviour [41]. Research revealed a number of cognitive abilities discussed in [41] by various authors.  These include general intellectual abilities, verbal, reading, second language, individual differences in learning and memory, mathematical, mental imagery, deductive and inductive reasoning and PS.

The idea *of thinking as an information-processing* was originated by Herbet Simon [10], who maintained that information transmission can be described in terms of behaviour of formally described symbol manipulation systems and combined it with the Gestalt psychology's ideas that long term memory is an organisation of different associations and that PS is goal directed.

Since Simon, many different theories emerged, but they all shared a certain common perspective.  Firstly, information-processing theories seek to explain mental events – the cognitive variables which intervene between stimulus and response.  Secondly, these cognitive events are best understood as symbolic representations and processes which act upon these mental representations in real time.  Thirdly, these representations and processes form the cognitive structure.  Finally comparisons (strong or weak versions) are drawn between the computations of the human mind and the computer [10].  This view of thinking has generated volumes of research programmes on cognitive phenomena such as perception, attention, memory, language, problem solving, reasoning and so on.

Information processing technologies aim at duplicating the mind (artificial intelligence) adding the advantage of operating at higher speeds than the mind in computations.  The complexity of such technologies rendered existing learning theories inadequate as more dimensions were added to learning.  Einhorn [38]  puts it very well when he states that *as we* *have changed technology, technology has also changed us, especially in how we think about thinking and seek new ways to solve the many questions and problems we face.*

While thinking can very from mundane to a very sophisticated and complex one and could be intentional or unintentional, a goal directed thinking is a kind of scientific thinking. And if computers are involved it becomes a kind of CT. CT has become a fundamental skill, ranking alongside reading, writing and arithmetic, it can be found on all the subjects [42]. But no real problem can be solved without an undisputed higher order thinking skill such as a critical thinking, which precedes any form of thinking skill in PS.

If we model a PS situation and see the problem as "an obstacle", then this creates two different hypothetical approaches to overcome this obstacle. These approaches elucidate the connection between critical and computational thinking, while the existing knowledge forms the link between them.

The first approach, presented graphically in Figure 1, contains a linear relationship between the above three constructs (processes) and PS is the product of them.  It does not feature in the processes explicitly.  The rationale is that each construct is a prerequisite to the next.

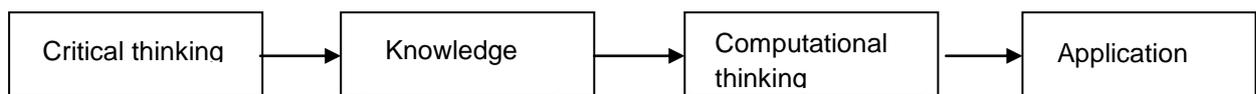

**Figure 1: Linear PS model [15]**

On the contrary, for the second approach, presented graphically in Figure 2, the various processes take place simultaneously.  The type of problem dictates the sequence of the relationships.  Again PS is the product of these processes.

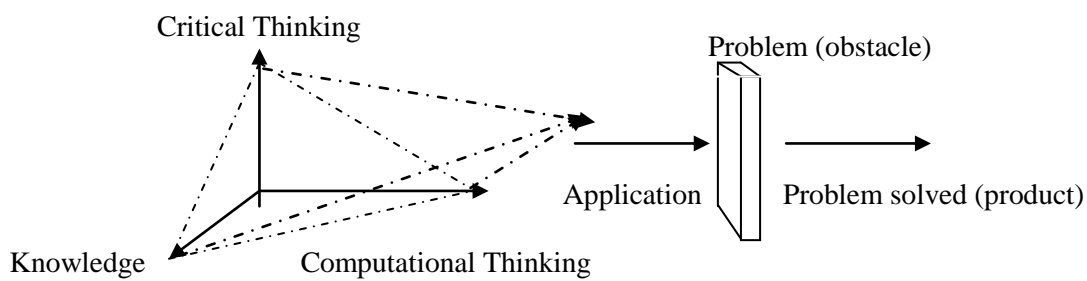

**Figure 2:  3-D problem solving model [15]**

These hypothetical models could be used to conceptualize PS of a real technological problem:
- For the linear PS model, once awareness of the problem is made the existing information is extracted and critically analyzed.  The problem solver then 'dips into' his/her knowledge base and thinking in a computer like scientific way applies that knowledge to solve the problem.

- In the 3-D PS model all three constructs act simultaneously and applied to the problem at hand.  If the problem is solved then the solution is the product of those cognitive actions.  However, this relationship among the constructs and subsequent application has to be justified.  Such a justification is based on the hypothesis that, if there exists sufficient background knowledge and through critical thinking the necessary new knowledge is retrieved, then computational thinking is applied and the problem is solved.

## 5. Classroom Experiments

Exploratory investigations have demonstrated how exposure to CT enhances the way students approach problems.  For example, Lewandowski et al. [43] illustrated the idea of "commonsense programming" for students without programming experience Students were asked to propose solutions to avoid selling theatre tickets for the same seat twice at multiple box offices.  The results showed that 69% of the solutions correctly identified a race condition, which indicated that the students were indeed equipped with a natural but underdeveloped understanding for solving problems computationally [44].

A team of mathematicians and mathematics educators led by Ed Dubinsky developed the APOS theory for teaching and learning mathematics [45] based on Piaget's principle that there is a close relationship between the nature of a mathematical concept and its development in the mind of an individual ([46], p. 13). According to APOS an individual deals with a mathematical situation by using certain mental mechanisms to build cognitive structures that are applied to the situation. The main mechanisms are called 'interiorization' and 'encapsulation' and the related structures are actions, processes, objects and schemas. The initial letters of the last four words constitute the acronym APOS. A basic component of the didactic approach of APOS, known as the 'ACE Teaching Cycle' [47], includes students' activities with computers (solution of real world problems, exercises, etc) that are properly designed to help them in developing the necessary cognitive structures. A series of quantitative and qualitative investigations, performed by Dubinsky's research team, proved the effectiveness of their method in a variety of mathematical topics including algebra, geometry, differential and integral calculus, discrete mathematics, etc  [47].

In our will to explore further the effect of the use of computers as a tool in solving mathematical problems we performed during the academic year 2011-12 the following classroom experiments:

In the first experiment the subjects were 90 students of the School of Technological Applications (prospective engineers) of the Graduate Technological Educational Institute (TEI) of Patras , Greece attending the course "Higher Mathematics I" [1] of their first term of studies. The students, who had no programming experience, where divided in two groups of 45 students in each of them.

In the (control) group the lectures were performed in the classical way on the board, followed by a number of exercises and examples connecting mathematics with real world applications and problems. The students participated in solving these problems.

The difference for the second (experimental) group was that part (about the 1/3) of the lectures and the exercises were performed in a computer laboratory. There the instructor used the suitable technological tools (computers, video projections, etc) to present the corresponding mathematical topics in a more "live" and attractive manner to students', while the students themselves, divided in small groups, used the existing ready mathematical software to solve the problems with the help of computers. Notice that all students (of both groups) were learning in a parallel course (Computer Science I) among the other basics about computers and the use of one of the well known mathematical software packages.

At the end of the term all students passed the final written examination of the mathematics course for the assessment of their progress. The examination involved a number of general theoretical questions and exercises covering all the topics taught and three simplified real world problems (see Appendix) requiring mathematical modeling techniques for their solution (time allowed was three hours). We marked the students' papers separately for the questions and exercises and separately for the problems.

In assessing the general performance of the two groups we applied the commonly GPA method [2]. used in the USA and other countries GPA. According to the marks obtained no significant differences were found for the two groups concerning the part of theoretical questions and exercises. On the contrary, the performance of the second (experimental) group was found to be significantly better in solving the problems (GPA$_1$ = 2, GPA$_2 \approx$ 2.49).

---

[1] The course involves Differential and Integral Calculus in one variable, Elementary Differential Equations and Linear Algebra.

[2] Te **G**reat **P**oint **A**verage (GPA) is a weighted average of the students' performance. For this, each student's paper is marked with A (90-100%), B (80-90%), C(80-70%), D (60-70%), or F (< 60%). Then, if $n$ is the total number of students and $n_A, n_B, n_C, n_D, n_F$ denote the numbers of students getting the marks A, B, C, D, F respectively, GPA = $\dfrac{0.n_F + 1.n_D + 2.n_C + 3n_B + 4.n_A}{n}$ .

At the same chronological period the same experiment was also performed under similar conditions with two groups of students of the School of Management and Economics of the TEI of Patras (100 students in each group). In this case the performance of the first (control) group was found to be slightly better for the first part of the examination (questions and exercises), but the performance of the experimental group was found again to be better for the second part (problems).

In concluding, the results of our experiments give a strong indication that the use of computers as a tool for PS enhances the students' abilities in solving real world mathematical problems.

## 6. Discussion and Conclusions

The above exploration into PS can be very simply summarised by the following phrases:

- *Solving:*  This is an active cognitive action by the problem solver.  Once he/she becomes aware of the problem, various cognitive processes (recall from memory, use of thinking skills and so on) come into being.

- *Computers:* A very useful tool for the development and improvement of students' PS skills.

- *Complex:*  The complexity of the problem requires analysis and synthesis, deciding on the best method, paying attention to detail and so on.  All these are *critical thinking* skills.

- *Technological problems:*  Technology requires a different approach to other fields since, if computers are involved, then one has to make the computer to 'think like them', but also encourage others 'think like a computer'.  All these are achieved through *computational thinking* that synthesises critical thinking and existing knowledge and applies them to solve the problem using computer science concepts and techniques.

- *Learning programming:* It becomes a demanding task in our nowadays society that requires CT to describe a complex problem and to propose a solution. And although students should become trained to CT before learning programming, the latter is probably the best way to learn CT explicitly.

- *Classroom experiments:* The results of our experiments, performed recently at the Graduate Technological Educational Institute of Patras in Greece, gave a strong indication that the use of computers as a tool for PS enhances the students' abilities in solving real world problems involving mathematical modelling. This is crossed by earlier findings of other researchers concerning the PS process in general (not only for mathematical problems).

The implication of these findings is very important to education as irrespective of the sequence that a problem is solved, critical thinking plays a central role in knowledge acquisition and creation, in computational thinking and thus in real complex technological problems.

However, we must underline a big danger hiding behind this reality. Indeed, people today using the convenient small calculators can make quickly and accurately all kinds of numerical operations. Further, the existence of a variety of suitable software mathematical packages gives the possibility of solving automatically all kinds of equations, to make any kind of algebraic operations, to calculate limits, derivatives, integrals, etc, and even more to obtain all the existing alternative proofs of the basic mathematical theorems and in some cases to produce new ones. Based on the above facts a number of scientists, mainly among the specialists of Computer Science, have already reached to the conclusion that teachers will not be needed in future for the development of students' knowledge base and learning skills, since everything could be done by the computers (possibly at home). "The use of horses is not necessary, from the time that cars were invented", argue some of them.

But, this is actually an illusion! In fact, the acquisition of information is valuable for the learner, but the most important thing is to learn how to think rationally and creatively. The latter is impossible to be succeeded through the help of computers only, because computers have been created by humans and, although they dramatically exceed in speed and memory, they will never reach, at least according to the standard logic, the quality of human thinking. On the other hand, the practice of students with numerical, algebraic and analytic calculations, with the solution of problems and the rediscovery of proofs of the known mathematical theorems, must be continued for ever; otherwise they will gradually loose the sense of numbers and symbols, the sense of space and time, and they will become unable to create new knowledge and technology.

We shall close by copying the following paragraph from the report of the Working Group on PS of the 10$^{th}$ International Conference of Mathematics Education into the 21$^{st}$ Century project [48]:

*"Future mathematics teaching has not only to focus on concepts and teaching techniques of computing, but also on PS and problem posing to reach general aims like creativity, ability of systematisation, abilities of communication, argumentation, presenting mathematics results, and ability of working in a team as well as getting a vivid view and a positive belief about mathematics and its application in real world".*

## References

[1] Wing, J. M. , "Computational thinking", *Communications of the ACM*, Vol.49,  33-35, 2006.

[2] King, P. M. &  Baxter-Magolda, M. B., "A developmental perspective on learning", *Journal of college student development*, 37(2), 163-173, 1996.

[3]  South African Qualifications Authority, "Regulations under the South African Qualifications Act, 1995" (Act No. 58 of 1995), *Government Gazette* No 6140, Vol. 393, No 18787, 28 March 1998, Pretoria, South Africa, 1998.

[4] Martinez, M., "What is meta cognition?  Teachers intuitively recognize the importance of meta cognition, but may not be aware of its many dimensions",  *Phi Delta Kappan,* 87(9), 696-714, 2007.

[5] Green, A. J. K. & Gillhooly, K. ,  "Problem solving".  In: Braisby, N.  & Gelatly, A. (Eds.), *Cognitive Psychology*, Oxford University Press,  Oxford, 2005.

[6] Schoenfeld, A. (Ed.), "Cognitive science and mathematics education", New Jersey: Erlbaum, 1987.

[7] Mc Peck, J. E.,  "Critical thinking and education", Martin Robinson, Oxford, 1981.

[8] Salomon, G. & Perkings, D., "Rocky roads to transfer: Rethinking mechanisms of a neglected phenomenon", *Educational Psychologist*, 24, 113-142, 1989.

[9] Voskoglou, M. G., "Problem Solving from Polya to Nowadays: A Review and Future Perspectives". In: R. V. Nata (Ed.), *Progress in Education*, Vol. 22, Chapter 4, 65-82, Nova Publishers, NY, 2011.

[10] Mc Guinness, C., "Teaching thinking: New signs for theories of cognition", *Educational Psychology,* 13(3-4), 305-316, 1993.

[11] Halpern, D., "Thought and knowledge: An introduction to critical thinking" (4[th] edition), Mahwah, NJ: Earlbaum, 2003.

[12] Matlin,  W. M., "Cognition", New York: Wiley & Sons, 2005.

[13] Polya, G., "How I solve it: A new aspect of mathematical method", New Jersey: Princeton University Press, 1973.

[14] Schoenfeld, A., "The wild, wild, wild, wild world of problem solving: A review of sorts", *For the Learning of Mathematics*, 3, 40-47, 1983.

[15] Giannakopoulos, A., "Problem solving in academic performance: A study into critical thinking and mathematics content as contributors to successful application of knowledge and subsequent academic performance", unpublished Ph.D. Thesis, University of Johannesburg, South Africa, 2012.

[16] Polya, G., "How to solve it", Princeton University Press, Princeton, 1945.

[17] Schoenfeld, A., "Teaching Problem Solving skills", *American Mathematical Monthly*, 87, 794-805, 1980.

[18] Lester F. K. , Garofalo,  J. & Kroll, D. L., "Self-confidence, interest, beliefs and meta cognition": Key influences on problem-solving behavior. In:  D. B. Mcleod & V. M. Adams (Eds.), *Affect and Mathematical Problem Solving: A New Perspective*, 75-88, Springer-Verlag, New York, 1989.

[19] Von Glaserseld, E., "Learning as a Constructive Activity". In: C. Janvier (Ed.), *Problems of representation in the teaching and learning of mathematics*, Lawrence Erlbaum, Hillsdale, N. J., 1987.

[20] Voskoglou, M. G., "The use of mathematical modelling as a learning tool of mathematics", *Quaderni di Ricerca in Didactica (Scienze Mathematche)*, 16, 53-60, Palermo, Italy, 2006.

[21] Carlson, M.  P. & Bloom, I., "The cyclic nature of problem solving: An emergent multidimensional problem-solving framework", *Educational Studies in Mathematics*, 58, 45-75, 2005.

[22] Sternberg, R. J.  & Ben-Zeev, T., "Complex cognition: The psychology of human thought",  Oxford University Press, Oxford, 2001.

[23] Schoenfeld, A., "How we think: A theory of goal-oriented decision making and its educational applications"*,* New York: Routledge, 2010.

[24] Pascarella,  E. T. & Terenzini,  P., "How college affects students", San Francisco: Jossey-Bass,  1991.

[25] Brookfield, S.D., "Developing critical thinkers: Challenging adults to explore alternative ways of thinking and acting", Open University Press, England, 1987.

[26] Williams, R. L., "Targeting critical thinking within teacher education: The potential impact on  Society", *The Teacher Educator,* 40(3),  163-187,  2005.

[27] Tiwari, A., Lai, P., So, M. & Yuen, K., "A comparison of the effects of problem-based learning and lecturing on the development of students' critical thinking", *Med. Educ.* ,41(2), 156-174, 2006.

[28] Papert,  S., "An exploration in the space of Mathematics Education", *International Journal of  Computers for Mathematics*, Vol. 1, No. 1, 95-123, 1996.

[29] Liu, J.  & Wang, L., "Computational Thinking in Discrete Mathematics", *IEEE 2$^{nd}$ International Workshop on Education Technology and Computer Science*, 413-416,  2010.

[30] Lu, J. J.  &  Fletcher, G. H. L., "Thinking about computational thinking", *Proceedings of the ACM Special Interest Group on Computer Science Education '09*, March 3-7, 260-264,  2009.

[31] Qualls, J. A.  & Sherrell, L. B., "Why computational thinking should be integrated into the curriculum", *Journal of Computing Sciences in Colleges*, 25, 66-71,  2010.

Egyptian Computer Science Journal ,ECS ,Vol.36 No.4, September 2012 ISSN-1110-2586[32] Anderson, R. C., Spiro, R. J. & Anderson, M. C., "Schemata as scaffolding for the representation of information in connected discourse", *American Educational Research Journal*, 15, 433-440, 1978.

[33] Sarjoughian, H. S. & Zeigler, B. P. , "Abstraction Mechanisms in discrete-event inductive modelling", *Proceedings of the Winter Simulation Conference*, 748-755, 1996.

[34] Barr, V. & Stephenson, C., "Bringing Computational Thinking to K-12: What is Involved and What is the Role of the Computer Science Education Community?, *ACM Inroads*, Vol. 2(1), 48-54, 2011.

[35] Magana, A. J., Marepalli, P. & Clark, J. V., "Work in Progress – Integrating Computational and Engineering Thinking through Online Design and Simulation of Multidisciplinary Systems", *41$^{st}$ ASEE/IEEE Frontiers in Education Conference*, October 12-15, 2011.

[36] Denning, P.J., "Beyond computational thinking", *Communications of the ACM*, 52(6), 28-30, 2009.

[37] Kazimoglu, C. , Kiernan, M., Bacon, L. & MacKinnon, L., "Understanding Computational Thinking Before Programming: Developing Guidelines for the Design of Games to Learn Introductory Programming Through Game-Play", *International Journal of Game-Based Learning*, 1(3), 30-52, 2011.

[38] Einhorn, S., "Micro-Worlds, Computational Thinking, and 21$^{st}$ Century Learning", *Logo Computer Systems Inc, White Paper*, 2012.

[39] Bennett, V., Koh, K. & Repenning, A., "Computer Science Education Re-Kindles Creativity in Public Schools", presented at the conference *ITCSE 2011*, Germany, 2011.

[40] Sarrazy B., "Paradoxes de la prise en compte des disparités culturelles et des individualités dans l'enseignement des mathématiques", *Proceedings CIEAEM 58*, 42-47, Srni, Czech Republic, 2006.

[41] Sternberg, R. J. (Ed.), "Beyond IQ: A triarchic theory of human intelligence", New York: Cambridge Press, 1985.

[42] Guzdial, M., "Paving the way for computational thinking", *Communications of the ACM,* 51(8), 25-27, 2008.

[43] Lewandowski, G., Bouvier, D., McCartney, R., Sanders, K. & Simon, B., "Common sense computing (episode 3): Concurrency and concert tickets", *Proceedings of the Third International Workshop on Computing Education Research (ICER '07),* 2007.

[44] Yadav, A., Zhou, N., Mayfield, C., Hambrusch, S. & Korb, J. T., "Introducing Computational Thinking in Education Courses", *Proceedings of the 42$^{nd}$ ACM technical symposium on Computer Science Education 11*, 465-470, March 9-12, 2011.
-44-

[45] Asiala, M., et al., "A framework for research and curriculum development in undergraduate mathematics education", *Research in Collegiate Mathematics Education II*, *CBMS Issues in Mathematics Education*, 6, 1-32, 1996.

[46] Piaget, J., "Genetic Epistemology", Columbia University Press, New York and London, 1970.

[47] Weller, K. et al., "Student performance and attitudes in courses based on APOS theory and the ACE teaching style". In A. Selden et al. (Eds.), *Research in collegiate mathematics education V*, 97-

   131, Providence, RI: American Mathematical Society, 2003.

[48] Graumann, G., "Working Group Report on Problem Solving", *10th International MEC21 Conference*, Dresden, Germany, 2009.

Available on line at http://math.unipa.it/~grim/21Project.htm/21Project_dresden_sept_2009.htm

## Appendix

*The problems given for solution to students in our classroom experiments:*

*Problem 1:* We want to construct a channel to run water by folding the two edges of an orthogonal metallic leaf having sides of length 20cm and 32 cm, in such a way that they will be perpendicular to the other parts of the leaf. Assuming that the flow of the water is constant, how we can run the maximum possible quantity of the water?

*Remark:* The correct solution is obtained by folding the edges of the longer side of the leaf. Some students solved the problem by folding the edges of the other side and failed to realize (validation of the model) that their solution was wrong.

*Problem 2:* Let us correspond to each letter the number showing its order into the alphabet (A=1, B=2, C=3 etc). Let us correspond also to each word consisting of 4 letters a 2X2 matrix in the obvious way; e.g. the matrix $\begin{bmatrix} 19 & 15 \\ 13 & 5 \end{bmatrix}$ corresponds to the word SOME. Using the matrix E=$\begin{bmatrix} 8 & 5 \\ 11 & 7 \end{bmatrix}$ as an encoding matrix how you could send the message LATE in the form of a camouflaged matrix to a receiver knowing the above process and how he (she) could decode your message?

*Problem 3:* The population of a country is increased proportionally. If the population is doubled in 50 years, in how many years it will be tripled?